\documentclass[a4paper,twoside]{article}

\usepackage{subfigure}
\usepackage{tabularx}
\usepackage[table]{xcolor}
\usepackage[normalem]{ulem}
\usepackage{algorithm}
\usepackage[section]{placeins}
\usepackage{array}

\usepackage{times}
\usepackage{graphicx}
\usepackage{xcolor}
\usepackage{pgfplots}
\pgfplotsset{compat=1.5}
\usepackage{caption}
\usepackage{multirow}
\usepackage{pifont}
\usepackage{hyperref}
\usepackage{ragged2e}
\usepackage{latexsym}

\definecolor{forestgreen}{rgb}{0.13, 0.55, 0.13}
\newcommand{\first}[1]{\textcolor{blue}{#1}}
\newcommand{\second}[1]{\textcolor{forestgreen}{#1}}
\newcommand{\third}[1]{\textcolor{red}{#1}}

\usepackage{epsfig}
\usepackage{calc}
\usepackage{amssymb}
\usepackage{amstext}
\usepackage{amsmath}
\usepackage{amsthm}
\usepackage{multicol}
\usepackage{pslatex}
\usepackage{apalike}
\usepackage{SCITEPRESS}     

\begin{document}

\title{Towards real-time object recognition and pose estimation in point clouds}

\author{\authorname{Marlon Marcon\sup{1}\orcidAuthor{0000-0001-9748-2824}, Olga Regina Pereira Bellon\sup{2}\orcidAuthor{0000-0003-2683-9704}, Luciano Silva\sup{2}\orcidAuthor{0000-0001-6341-1323}}
\affiliation{\sup{1}Dapartment of Software Engineering, Federal University of Technology - Paran\'{a}, Dois Vizinhos, Brazil}
\affiliation{\sup{2}Department of Computer Science, Federal University of Paran\'{a}, Curitiba, Brazil}
\email{marlonmarcon@utfpr.edu.br, olga@ufpr.br, luciano@inf.ufpr.br}
}

\keywords{Transfer Learning, 3D Computer Vision, Feature-based Registration, ICP Dense Registration, RGB-D Images.}

\abstract{Object recognition and 6DoF pose estimation are quite challenging tasks in computer vision applications. Despite efficiency in such tasks, standard methods deliver far from real-time processing rates. This paper presents a novel pipeline to estimate a fine 6DoF pose of objects, applied to realistic scenarios in real-time. We split our proposal into three main parts. Firstly, a Color feature classification leverages the use of pre-trained CNN color features trained on the ImageNet for object detection. A Feature-based registration module conducts a coarse pose estimation, and finally, a Fine-adjustment step performs an ICP-based dense registration. Our proposal achieves, in the best case, an accuracy performance of almost 83\% on the RGB-D Scenes dataset. Regarding processing time, the object detection task is done at a frame processing rate up to 90 FPS, and the pose estimation at almost 14 FPS in a full execution strategy. We discuss that due to the proposal's modularity, we could let the full execution occurs only when necessary and perform a scheduled execution that unlocks real-time processing, even for multitask situations.}

\onecolumn \maketitle \normalsize \setcounter{footnote}{0} \vfill

\section{\uppercase{Introduction}}
\label{sec:introduction}

\noindent Object recognition and 6D pose estimation represent a central role in a broad spectrum of computer vision applications, such as object grasping and manipulation, bin picking tasks, and industrial assemblies verification \cite{vock2019fast}. Successful object recognition, highly reliable pose estimation, and near real-time operation are essential capabilities and current challenges for robot perception systems.

A methodology usually employed to estimate rigid transformations between scenes and objects is centered on a feature-based template matching approach.  Assuming we have a known item or a part of an object, this technique involves searching all the occurrences in a larger and usually cluttered scene \cite{vock2019fast}. However,  due to natural occlusions, such occurrences may be represented only by a partial view of an object. The template is often another point cloud, and the main challenge of the template matching approach is to maintain the runtime feasibility and preserve the robustness. 

Template matching approaches rely on RANSAC-based feature matching algorithms, following the pipeline proposed by \cite{aldoma2012tutorial}. RANSAC algorithm has proven to be one of the most versatile and robust. Unfortunately, for large or dense point clouds, its runtime becomes a significant limitation in several of the example applications mentioned above \cite{vock2019fast}. When we seek a 6Dof estimation pose, the real-time is a more challenging task \cite{marcon2019boosting}. In an extensive benchmark of full cloud object detection and pose estimation, \cite{hodan2018bop} reported runtime of a second per test target on average.

Deep learning strategies for object recognition and classification problems have been extensively studied for RGB images. As the demand for good quality labeled data increases, large datasets are becoming available, serving as a significant benchmark of methods (deep or not) and as training data for real applications. ImageNet \cite{deng2009imagenet} is, undoubtedly, the most studied dataset and the \textit{de-facto} standard on such recognition tasks. This dataset presents more than 20,000 categories, but a subset with 1,000 categories, known as ImageNet Large Scale Visual Recognition Challenge (ILSVRC), is mostly used.

Training a model on ImageNet is quite a challenging task in terms of computational resources and time consumption. Fortunately, transferring its models offer efficient solutions in different contexts, acting as a blackbox feature extractor. Studies like  \cite{agrawal2014analyzing} explore and corroborate this high capacity of transferring such models to different contexts and applications. Regarding the use of pre-trained CNN features, some approaches handle the object recognition on the Washington RGB-D Object dataset, e.g., \cite{zia2017rgb} with the VGG architecture and \cite{caglayan2020cnns} evaluate several popular deep networks, such as AlexNet, VGG, ResNet, and DenseNet. 

This paper introduces a novel pipeline to deal with point cloud pose estimation in uncontrolled environments and cluttered scenes. Our proposed pipeline recognizes the object using color feature descriptors, crops the selected bounding-box reducing the scenes' searching surface, and finally estimates the object's pose in a traditional local feature-based approach. Despite adopting well-known techniques in the 2D/3D computer vision field, our proposal's novelty centers on the smooth integration between 2D and 3D methods to provide a solution efficient in terms of accuracy and time. 

\section{\uppercase{Background}}

Recognition systems work with objects, which are digital representations of tangible real-world items that exist physically in a scene. Such systems are unavoidably machine-learning-based approaches that use features to locate and identify objects in a scene reliably. Together with the recognition, another task is to estimate the location and orientation of the detected items. In a 3D world, we estimate six degrees of freedom (6DoF), which refers to the geometrical transformation representing a rigid body's movement in a 3D space, i.e., the combination of translation and rotation.

\subsection{Color feature extraction}

As a mark on the deep learning history, \cite{krizhevsky2012alexnet} presented the first Deep Convolutional Architecture employed on the ILSVRC, an 8-layer architecture dubbed AlexNet. This network was the first to prove that deep learning could beat hand-crafted methods when trained on a large scale dataset. After that, ConvNets became more accurate, deeper, and bigger in terms of parameters. \cite{simonyan2014vgg} propose VGG, a network that doubled the depth of AlexNet, but exploring tiny filters ($3 \times 3$), and became the runner-up on the ILSVRC, one step back the GoogLeNet \cite{szegedy2015googlenet}, with 22 layers. GoogLeNet relies on the Inception architecture \cite{szegedy2016inception}. Another type of ConvNets, called ResNets \cite{he2016resnet}, uses the concept of residual blocks that use skip-connection blocks that learn residual functions regarding the input. Many architectures have been proposed based on these findings, such as ResNet with 50, 101, and 152 \cite{he2016resnet}. Also, based on developments regarding the residual blocks, \cite{xie2017resnext} developed the ResNeXt architecture. The basis upon ResNeXt blocks resides on parallel ResNet-like blocks, which have the output summed before the residual calculation. Some architectures propose the use of Deep Learning features on resource-limited devices, such as smartphones and embedded systems. The most prominent architecture is the MobileNet \cite{sandler2018mobilenetv2}. Another family of leading networks is the EfficientNet \cite{tan2019efficientnet}. Relying on the use of these lighter architectures, EfficientNet proposes very deep architectures without compromise resource efficiency.

\subsection{Pose estimation}

As presented in \cite{aldoma2012tutorial}, a comprehensive registration process usually consists of two steps: coarse and fine registrations.  We can produce a coarse registration transformation by performing a manual alignment, motion tracking or, the most common, by using the local feature matching. Local-feature-matching-based algorithms automatically obtain corresponding points from two or multiple point-clouds, coarsely registering by minimizing the distance between them. These methods have been extensively studied and have confirmed to be compliant and computer efficient \cite{guo2016comprehensive}. After coarsely registering the point clouds, a fine-registration algorithm is applied to refine the initial coarse registration iteratively. Examples of fine-registration algorithms include the ICP algorithm that perform point-to-point alignment \cite{besl1992icp}, or point-to-plane \cite{chen1992icp}. These algorithms are suitable for matching between point-clouds of isolated scenes (3D registration) or between a scene and a model (3D object recognition). This proposal adopted two approaches to generate the initial alignment: a traditional feature-based RANSAC and the Fast Global Registration (FGR) \cite{zhou2016fast}. 

\section{\uppercase{Proposed approach}}

\begin{figure*}[ht]
	\centering
	\includegraphics[width=0.94\linewidth]{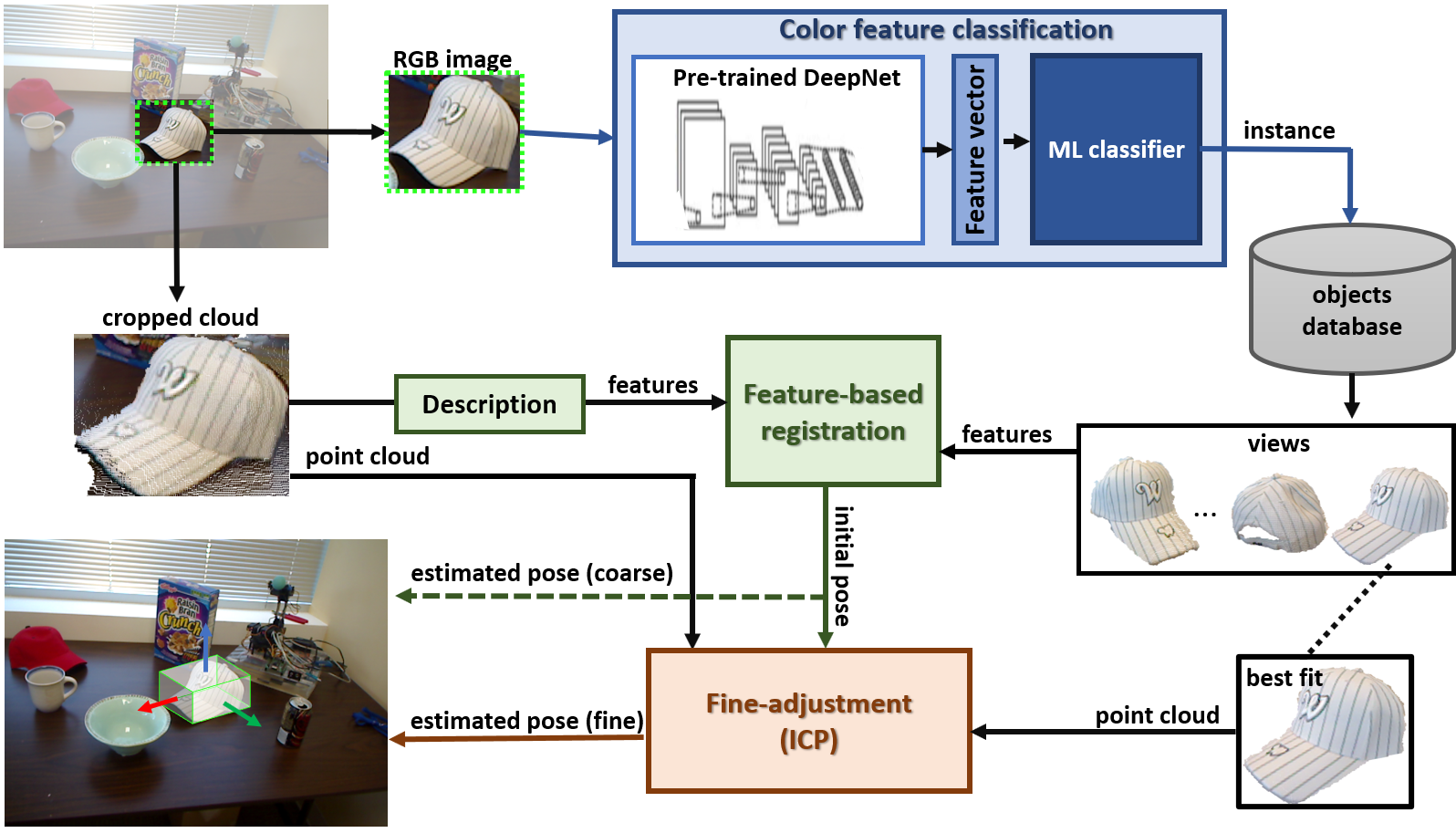}
	\caption[Pipeline of the proposed approach to pose estimation]{Pipeline of the proposed approach to pose estimation. To estimate the pre-segmented object's instance, we extract its features by a deep learning color-based extractor and a pre-trained ML classifier. After selecting the objects database, the view with the highest number of correspondences resulting from a feature-based registration algorithm. Finally, we apply an ICP dense registration algorithm to estimate the position and pose of the object.}
	\label{fig:pipeline-application}
\end{figure*}

In this section, we explain in detail our proposed approach. Our proposed pipeline starts from an RGB image and its corresponding point cloud, generated from RGB and depth images. These inputs are submitted to our three-stage architecture: color feature classification, feature-based registration, and fine adjustment. We depict our proposal in \autoref{fig:pipeline-application} and present these steps in the next sections.

\noindent 

\subsection{Color feature classification}
\label{subsec:color-feature-classification}

Our proposal starts detecting the target object and estimating a bounding box of it. After this detection, we preprocess the image and submit to a deep-learning-based color feature extractor. The preprocessing step includes image cropping and resizing to adjust to the network input dimensions. The deep network architectures employed in our experiments output a feature vector, 1000 bins long, used to predict the object's instance, by a pre-trained ML classifier. We emphasize that our approach is size-independent regarding the feature vector, but for a fair comparison we chose networks with the same output size.

In our trials, we explored the achievements of \autoref{tab:cnn_color_rgbd}, and selected the most accurate networks: ResNet101 \cite{he2016resnet}, MobileNet v2 \cite{sandler2018mobilenetv2}, ResNeXt101 32$\times$8d \cite{xie2017resnext}, and EfficientNet-B7 \cite{tan2019efficientnet}. These networks input a $224\times244$ pixel image and output a 1000 bins feature vector. We employed the Logistic regression classifier, chosen after a performance evaluation of standard classifiers, to name: Support Vector Classifier (SVC) with linear and radial-based kernels, Random forest, Multilayer perceptron, and Gaussian naïve Bayes. We explore two variants of our ML model: a pre-trained on the Washington RGB-D Object dataset, and a distinct model, also in such dataset, but with a reduced number of objects, i.e., those annotated on the Washington RGB-D Scenes dataset. The latter provides an application-oriented approach, reducing the number of achievable classes, the inference time, and model size (\autoref{tab:rgbd-Scenes-full-vs-scenes}). To verify the best accurate classifier, we do not perform object detection. Instead, we get the ground-truth bounding boxes provided by the dataset, hence verifying for each ML system which is the best feasible performance.

\subsection{Feature-based registration}

We build a model database by extracting and storing useful information about the objects in a previous step. The database is composed of information concerning each item, as well as the extracted features of them. We choose a local-descriptors-based approach to estimate the object's pose. For each instance of an object, we store several partial views of it. Between these views, our method will select the most likely to the correspondent object on the scene.

Based on the predicted objects' classes, we can select a set of described views from the models' database. We then perform a feature-based registration between these views and the point cloud of the scene's object (previously cropped based on the detected bounding box). This method will estimate a transformation based on the correspondences between a scene and a partial view of an object. Then, the view with the highest number of inliers and at least three correspondences is selected. The estimated affine transformation will be input to the ICP algorithm and perform a dense registration.

We process each cloud with a uniform sampling as a keypoint extractor, adopting a leaf size of 1 cm. After, we describe each keypoint using the FPFH \cite{rusu2009fast} descriptor with a radius of 5 cm. We choose this descriptor due to its processing time and size (33 bins), well-suited for real-time applications. Methods like CSHOT \cite{salti2014shot} describes the color and geometric information and has proven to be an accurate solution applied to single object recognition on RGB-D Object dataset \cite{ouadiay2016}. However, with a descriptor length of 1344 bins, it is not suitable for real-time feature-matching. Another proposal that deals with color is PFHRGB \cite{rusu2008pfh}, which, despite being shorter (250 bins) than CSHOT, presents inefficient calculation time \cite{marcon2019boosting}. 

To perform the coarse registration step, we test two methods previously presented: RANSAC and FGR. We considered for both techniques an inlier correspondence distance lower than 1 cm between scene and models. We set the convergence criteria for RANSAC to 4M iterations and 500 validation steps, and for FGR to 100 iterations, following \cite{choi2015robust} and \cite{zhou2016fast}.

\subsection{Fine-adjustment}

The previous step outputs an affine transformation that could work as a final pose of the object concerning the scene. However, to guarantee a fine-adjustment, we employ an additional step to the process. We adopt the ICP algorithm, based on the point-to-plane approach \cite{chen1992icp}, to perform a dense registration. We use the transformation resultant from the registration step, the scene, and best-fitted view clouds as input. We set the maximum correspondence distance threshold to 1 cm. It is important to point that again, our proposal is generic, and the fine adjustment algorithm employed in this stage is flexible. Methods such as ICP point-to-point \cite{besl1992icp} and ColoredICP \cite{park2017colored} are perfectly adapted to our pipeline.

\section{\uppercase{Experimental Results}}

\subsection{Dataset}

\begin{figure}[htb]
	\centering
	\includegraphics[clip, trim=0.8cm 0cm 0.8cm 0cm,width=1\linewidth]{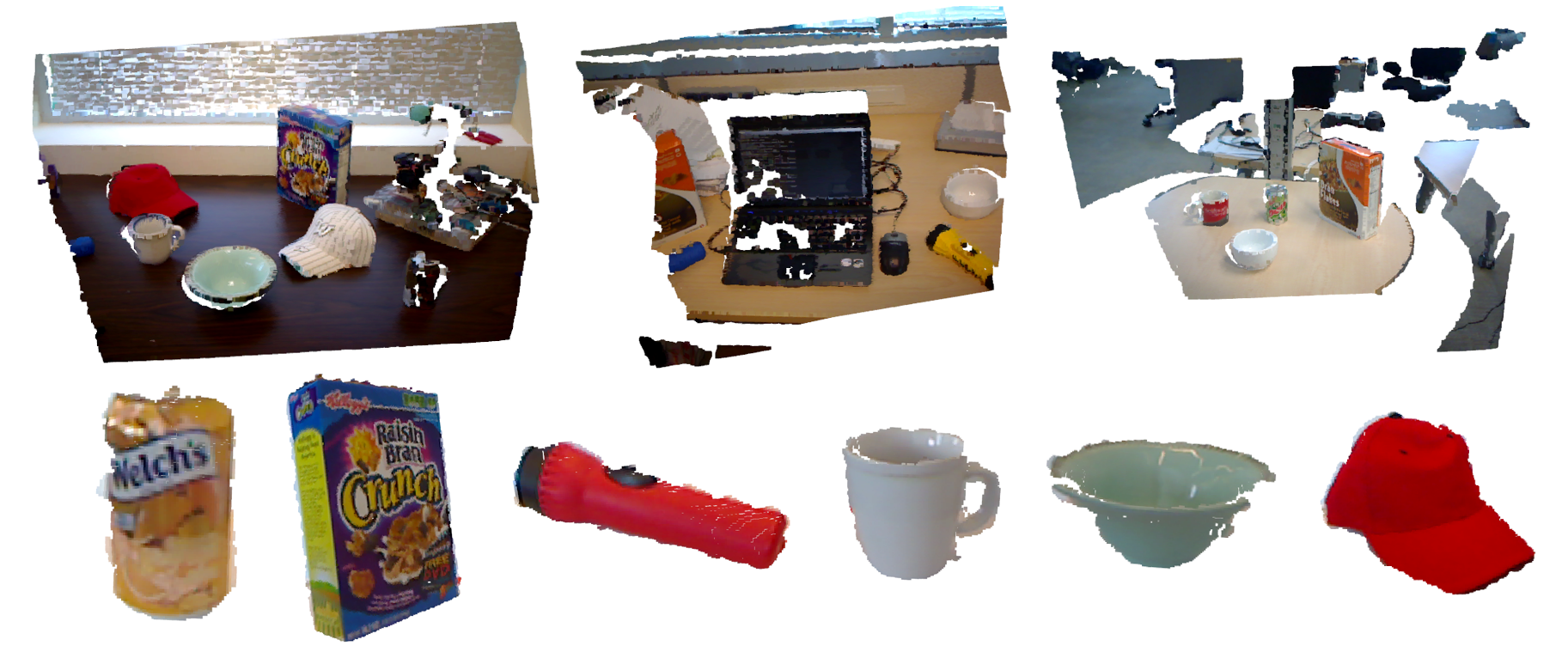}
	\caption{Examples of models and scenes from the Washington RGB-D Scenes dataset (top row), and objects from the RGB-D Object dataset (bottom row). Source: Adapted from \cite{lai2011rgbd}.}
	\label{fig:rgbd-scenes}
\end{figure}

We validate our proposal on the Washington RGB-D Object and Scenes datasets. Proposed by \cite{lai2011rgbd} the RGB-D Object contains a collection of 300 instances of household objects, grouped in 51 distinct categories. Each object includes a set of views, captured from different viewpoints with a Kinect sensor. A collection of 3 images, including RGB, depth, and mask is presented for each view. In total, this dataset has about 250 thousand distinct images. The authors also provide a dataset of scenes, named RGB-D Scenes. This evaluation dataset has eight video sequences of every-day environments. A Kinect sensor positioned at a human eye-level height acquires all the images at a  $640 \times 480$ resolution. This dataset is related to the first one, composed of 13 of the 51 object categories on the Object dataset. These objects are positioned over tables, desks, and kitchen surfaces, cluttered with viewpoints and occlusion variation, and have annotation at category and instance levels. A bidimensional bounding box represents the ground-truth of each object's position. \autoref{fig:rgbd-scenes} presents examples of both datasets. \autoref{tab:rgbd-Scenes-details} gives some details regarding the name and size of the sequences, and their average number of objects.

\begin{table}[htb]
	\centering
	\small
	\vspace{2.5mm}
	\setlength{\tabcolsep}{2pt}
	\renewcommand{\arraystretch}{1.3}
	\caption{Details regarding the RGB-D Scenes datasets. \label{tab:rgbd-Scenes-details}}
	\begin{tabular}{lcc}
		\hline
		Scene & Number of frames & Models per frame \\ \hline
		desk\_1 & 98 & 1.89 \\
		desk\_2 & 190 & 1.85 \\
		desk\_3 & 228 & 2.56 \\
		kitchen\_small\_1 & 180 & 3.55 \\
		meeting\_small\_1 & 180 & 8.79 \\
		table\_1 & 125 & 5.92 \\
		table\_small\_1 & 199 & 3.68 \\
		table\_small\_2 & 234 & 2.89 \\ \hline
		{Average} & {179.25} & {3.89} \\ \hline
	\end{tabular}
	\vspace{2mm}
\end{table}

\subsection{Evaluation Protocol}

We evaluate our proposal, quantitatively, and qualitatively. First, we consider CNN feature extraction and classification accuracy based on the models trained in the Object dataset (\autoref{tab:cnn_color_rgbd}). We also verify the entire dataset's processing time, looking at the frame processing rate in both classification and pose estimation scenarios.

As the Scenes dataset does not provide ground-truth annotations concerning the objects' pose, we had to find a plausible metric to evaluate the registration results. We adopted two different metrics: the Root mean squared error (RMSE) and an inlier ratio measurement. The latter represents the overlapping area between the source (model) and the target (scene). It is calculated based on the ratio between inlier correspondences and the number of points on the target. We also evaluate the correctness of predictions, both of object presence and pose. To do so, we follow \cite{marcon2019boosting} and employ the Intersection over Union metric (IoU), defined as:

\begin{equation} \label{eq:IoU}
IoU = \frac{BB_{GT} \cap BB_{Est}}{BB_{GT} \cup BB_{Est}}
\end{equation}

\noindent we consider $BB_{GT}$ the 3D projection of the 2D bounding box, provided as ground-truth. $BB_{Est}$ refers to the 3D bounding box that circumscribes the selected object view after applying the resulting transformation. 

We found experimentally that, for this particular dataset, when we estimate the IoU between the object 3D BB and the scene 2D projection, often the former is fully contained in the latter. However, due to their sizes, the calculated IoU is too low. Hence, we consider another metric, which we call Model Intersection Ratio (MIR) that represent the intersection volume over the model estimation volume:

\begin{equation} \label{eq:MIR}
MIR = \frac{BB_{GT} \cap BB_{Est}}{BB_{Est}}
\end{equation}

With the MIR metric, we guarantee that despite the IoU, when the estimation transform places the object inside (or nearly inside) the ground-truth 3D projection, a successful detection is performed. We consider a true positive when the $IoU \geq 0.25$ or the $MIR \geq 0.90$.

We compared our proposal with the standard 3D object recognition and pose estimation pipeline \cite{aldoma2012tutorial}, and with a boosted version of such pipeline, proposed by \cite{marcon2019boosting}. To calculate precision-recall curves (PRC), we varied the threshold on the minimum geometrically consistent correspondences, starting from at least three, related to each object's best-suited partial view. The area under the PRC curve (AUC) is then calculated and provides comparative results that assess our proposals' efficiency against traditional approaches.

\subsection{Implementation details}

We performed our tests on a Linux Ubuntu 18.04 LTS machine, equipped with a CPU Ryzen 7 2700X, 32GB of RAM, and a GPU Geforce RTX 2070 Super. To process the point clouds, perform keypoint extraction, description with FPFH, and registration with RANSAC and FGR, we used the Open3D Library. We preprocess images using Pillow and OpenCV. Deep learning models were implemented in PyTorch, and the pre-trained models extracted from torchvision. To implement traditional and boosted versions of object recognition and pose estimation pipelines, we use PCL 1.8.1, OpenCV 3.4.2, and the saliency detection of \cite{hou2017deeply}, following \cite{marcon2019boosting}.

\subsection{Results}

This section summarizes the Washington RGB-D Scenes' experimental evaluation results in terms of accuracy and processing time. 

\subsubsection{Object detection benchmark}

To assess the generalization capacity of CNN pre-trained models, we perform an object detection evaluation on the Object dataset \cite{lai2011rgbd}. \autoref{tab:cnn_color_rgbd} present results regarding classification of partial views of objects. We evaluate the instance recognition scenario, following \cite{lai2011rgbd}, i.e., considering Alternating contiguous frame (ACF) and Leave-sequence-out (LSO) scenarios. We compared our results with state-of-the-art object detection methods on this dataset. We perceived that pre-trained networks provide reliable results as off-the-shelf color feature extractors. In both evaluation approaches, tested networks present competitive results concerning the other competitors. In LSO, ResNet101 \cite{he2016resnet} features figures in the third position, and in ACF, 5 of 7 architectures outperform previous proposals.

\begin{table}[htb]
	\centering
	\small
	\vspace{2.5mm}
	\setlength{\tabcolsep}{2pt}
	\renewcommand{\arraystretch}{1.3}
	\caption{Comparison of CNN color features on the Washington RGB-D Object dataset. The best result reported in \first{blue}, the second best in \second{green}, and the third in \third{red}. \label{tab:cnn_color_rgbd}}
	\resizebox{0.48\textwidth}{!}{
	\begin{tabular}{lccc}
		\hline
		{Method} 					& LSO & ACF \\ \hline
		Lai \textit{et al.} (RF) \cite{lai2011rgbd} & 59.9 & 90.1 $\pm$ 0.8 \\
	    Lai \textit{et al.} (kSVC) \cite{lai2011rgbd} & 60.7 & 91.0 $\pm$ 0.5 \\
	    IDL \cite{lai2011sparse} & - & 54.8 $\pm$ 0.6 \\
	    SP+HMP \cite{bo2013unsupervised} & 92.1 &  - \\
	    Multi-Modal \cite{schwarz2015rgb} & 92.0 & - \\
	    MDSI-CNN \cite{asif2017multi} & \first{97.7} & - \\
	    MM-LRF-ELM \cite{liu2018multi} & 91.0 & - \\
	    HP-CNN \cite{zaki2019viewpoint} & \second{95.5} & - \\ \hline
		{AlexNet} \cite{krizhevsky2012alexnet} &  89.8 	&  93.9 $\pm$ 0.4              \\
		{ResNet101}\cite{he2016resnet}	&  \third{94.1} 	&  95.3 $\pm$ 0.3  \\
		{VGG16} \cite{simonyan2014vgg} 	&  88.8 	& 91.0 $\pm$ 0.6                    \\
		{Inception v3} \cite{szegedy2016inception} 	&  88.1 	&  90.3 $\pm$ 0.4     \\
		{MobileNet v2} \cite{sandler2018mobilenetv2}	&  93.8	&  \first{95.8 $\pm$ 0.3}    \\
		{ResNeXt101 $32\times 8$d} \cite{xie2017resnext}&  93.9 &  \second{95.7 $\pm$ 0.4}   \\ 
		{EfficientNet B7} \cite{tan2019efficientnet}	&  93.8 &  \third{95.6 $\pm$ 0.5}                            \\\hline
	\end{tabular}
	}
\end{table}

\begin{table*}[hbt]
	\centering
	\small
	\vspace{2.5mm}
	\renewcommand{\arraystretch}{1.3}
	\caption[Instance classification performance on the RGB-D Scenes datasets]{Instance classification performance on the RGB-D Scenes datasets. \label{tab:rgbd-Scenes-instance-application}}
\begin{tabular}{l|cc|cc|cc|cc}
\hline
 & \multicolumn{2}{c|}{MobileNet v2} & \multicolumn{2}{c|}{Resnet101} & \multicolumn{2}{c|}{ResNeXt101 32x8d} & \multicolumn{2}{c|}{EfficientNet-B7} \\ \cline{2-9} 
{Scene} & {Acc} & {FPS} & {Acc} & {FPS} & {Acc} & {FPS} & {Acc} & {FPS} \\ \hline
desk\_1 & 42.70\% & {13.03} & {51.89\%} & 9.66 & 48.11\% & 7.63 & 49.73\% & 6.55 \\
desk\_2 & 41.76\% & {12.95} & 38.92\% & 9.31 & 55.40\% & 7.93 & {76.42\%} & 6.35 \\
desk\_3 & 72.77\% & {9.84} & 52.57\% & 7.09 & 52.91\% & 5.78 & {90.58\%} & 4.60 \\
kitchen\_small\_1 & 36.31\% & {7.97} & 34.74\% & 5.29 & 48.20\% & 4.12 & {56.81\%} & 3.25 \\
meeting\_small\_1 & 41.40\% & {3.29} & 38.05\% & 2.35 & 42.92\% & 1.74 & {50.63\%} & 1.33 \\
table\_1 & 56.76\% & {4.62} & 38.11\% & 3.43 & 31.08\% & 2.49 & {61.49\%} & 2.00 \\
table\_small\_1 & 75.03\% & {7.50} & 63.30\% & 5.33 & 65.35\% & 3.89 & {83.36\%} & 3.16 \\
table\_small\_2 & 55.39\% & {9.13} & 45.35\% & 6.88 & 49.34\% & 5.04 & {65.88\%} & 4.10 \\ \hline
{Average} & 52.77\% & {6.99} & 45.37\% & 5.03 & 49.16\% & 3.80 & {66.86\%} & 3.02 \\ \hline
\end{tabular}
\end{table*}

Despite the significant results, this evaluation is essential to select the most suitable to perform object recognition in realistic scenarios, such as those presented by the Scenes dataset \cite{lai2011rgbd}. As the trials' output, we selected the top-four architectures to apply in our proposed pipeline.

\subsubsection{Object recognition in real-world scenes}
We opposed the selected CNN architectures examining only a classification based on the RGB information, taking the annotated bounding box, and submitting to the \textit{Color Feature Classification} stage of our pipeline (as in Section \ref{subsec:color-feature-classification}). \autoref{tab:rgbd-Scenes-instance-application} relates to instance-level recognition.

The first outcome of this evaluation is the dominance of two networks over the other competitors considering different aspects. EfficientNet \cite{tan2019efficientnet} architecture outperforms in terms of accuracy, and MobileNet v2 \cite{sandler2018mobilenetv2} in terms of processing time w.r.t. the others in almost all scenes. 

EfficientNet reaches an average accuracy of almost 67\%, followed by MobileNet v2, with almost 53\%. However, when we aim efficiency in processing time, EfficientNet does not perform so well, being the slowest network with a frame-rate of 3.02 per second. On the other hand, the MobileNet v2 fulfills the network's main proposal to be time-efficient and accurate for embedded applications. It presents the second-best accuracy and the best frame-rate, with almost 7 FPS.

The full-set of the Object dataset contains 51 categories and 300 distinct instances. Concerning the Scenes dataset, the number of annotated samples drops to 6 categories and 22 instances, i.e., only a small set of objects of Object dataset is achievable on the Scenes dataset. When we use a model trained on the full-set, most categories or instances will never be detected. Thou, we learned a lighter classifier that considers only such specific instances (\autoref{tab:rgbd-Scenes-full-vs-scenes}).

\begin{table}[htb]
	\centering
	\small
	\setlength{\tabcolsep}{5pt}
	\renewcommand{\arraystretch}{1.3}
	\caption{Performance comparison between full and a specific training set with objects from the Scenes dataset. \label{tab:rgbd-Scenes-full-vs-scenes}}
	\begin{tabular}{l|cc|cc}
		\hline
\multirow{2}{*}{DeepNet} & \multicolumn{2}{c|}{Full} & \multicolumn{2}{c}{Scenes} \\ \cline{2-5} 
 & Acc & \multicolumn{1}{c|}{FPS} & Acc & \multicolumn{1}{c}{FPS} \\ \hline
		MobileNet v2 & 52.77\% & {6.99} & 67.35\% & {24.62} \\
		Resnet101  &  45.37\% & 5.03 & 61.41\% & 13.94 \\
		ResNeXt101 32x8d & 49.16\% & 3.80 & 59.04\% & 8.86 \\
		EfficientNet-B7  & {66.86\%} & 3.02 & {82.94\%} & 5.88 \\ \hline
	\end{tabular}
\end{table}

After this change on the model specificity, we distinguish a noticeable improvement in accuracy and the processing time, achieving MobileNet v2 a near real-time performance on average. A significant gain on accuracy was established, with over 10\% for every architecture, pulling the best result to 83\% for EfficientNet.

Regarding the frame processing rate, it is essential to notice that the average number of models varies from 1.85 to 8.79 over the scenes, with almost four objects per frame in mean (\autoref{tab:rgbd-Scenes-instance-application}). Thus, we can infer that our proposal can deliver a near-real-time FPS, inclusive in a multi-classification problem. When we consider only a single target, the performance is almost four times faster, as presented in \autoref{tab:comparison-fast-ransac}, on the \textit{Color only} column.

\subsubsection{Pose estimation results}

Based on the assumption that we mapped the objects we aim to detect in a real-world scenario, we adopted those models trained on the RGB-D Object dataset subset. We considered only the instance detection situation. The reason for disregarding categories is that we could have intra-class misclassifications, corrupting the pose alignment step. For each instance detected by the \textit{Color feature classification} stage, we take ten views of the referred object from the models' database.

\begin{table*}[ht]
	\centering
	\small
	\vspace{2.5mm}
	\renewcommand{\arraystretch}{1.3}
	\caption{Comparison between feature-based registration methods. Values reported consider the processing time (in seconds) for ten views of the same object and the ICP for the best one selected. \label{tab:ransac-vs-fast-quantitative}}
	\begin{tabular}{lcccc}
		\hline
		\multicolumn{1}{c}{{Methods}} & \begin{tabular}[c]{@{}c@{}}{Feature-based} \\ {time} ($\downarrow$)\end{tabular} & {ICP time} ($\downarrow$) & {Inlier ratio} ($\uparrow$) & {RMSE} ($\downarrow$) \\ \hline
		RANSAC & 0.7688 & {0.0061} & {0.2689} & {0.0055} \\
		FGR \cite{zhou2016fast} & {0.0580} & 0.0075 & 0.1895 & 0.0059 \\ \hline
	\end{tabular}
\end{table*}

In \autoref{tab:ransac-vs-fast-quantitative} we report an evaluation concerning the Feature-based registration and Fine-adjustment stages of our pipeline. Getting a set of ten randomly selected views of the same object, we perform a coarse estimation by using RANSAC or FGR. We evaluate quantitatively such methods concerning the inlier ratio, RMSE, and execution time. We apply the resulting affine transformation as the input of an ICP dense registration and evaluate if this input can imply differences in the processing time.

Indeed, the FGR method is much faster than RANSAC. However, we observe that for both metrics RANSAC outperforms it. The Inlier ratio presented by the latter is around 50\% higher than the faster method and also shows an RMSE more consistent.  The transformation generated by the coarse alignment algorithm also impacts the ICP execution and we notice that a better estimation can speed up the fine-adjustment process. 

To evaluate more deeply if the ICP, after the feature-matching application, can surpass problems like a more rough estimation, we must assess an annotated pose. Unfortunately, the adopted dataset does not offer such data, and further studies may verify that affirmation on a pose-annotated dataset. However, we can evaluate the estimation correctness by employing the IoU and MIR metrics and verify if the feature-based registration step's estimation is reliable compared to standard approaches. In \autoref{tab:comparison-auc-fps} we perform such comparison regarding the AUC and FPS values of different setup of our proposed pipeline, the standard \cite{aldoma2012tutorial}, and the boosted \cite{marcon2019boosting} pipelines.

\vspace{3mm}
\begin{table}[htb]
\centering
\small
\renewcommand{\arraystretch}{1.3}
	\caption{Comparison of the proposed pipeline with standard object recognition and pose estimation approaches. Baseline refer to \cite{aldoma2012pcl} and Boost to \cite{marcon2019boosting}. Every trial employed FPFH \cite{rusu2009fast} as local descriptor with a uniform sampling as keypoint detector. Excepting the fisrt two rows, leaf size was set to 1 cm.. \label{tab:comparison-auc-fps}}
\begin{tabular}{lcc}
\hline
Method & AUC & FPS \\ \hline
Baseline $US_{0.02}$ & 0.0401 & 0.0023 \\ 
Boost $US_{0.02}$ & 0.0868 & 0.0918 \\ 
Boost $US_{0.01}$ & 0.1372 & 0.0339\\ \hline 
Resnet101 + FGR & 0.2228 & 13.8321 \\
ResNet101 + RANSAC & 0.2092 & 1.9649 \\
MobileNet v2 + FGR & 0.2922 & 13.8939 \\
MobileNet v2 + RANSAC & 0.2781 & 1.8905 \\
ResNeXt101 32x8d + FGR & 0.2090 & 14.1813 \\
ResNeXt101 32x8d + RANSAC & 0.1947 & 2.0268 \\
EfficientNet-B7 + FGR & 0.4123 & 8.9429 \\
EfficientNet-B7 + RANSAC & 0.2994 & 1.4344 \\ \hline
\end{tabular}
\end{table}

Results of \autoref{tab:comparison-auc-fps} confirm our claim that performing the object detection on the RGB images improves results compared to traditional approaches. Both standard and boosted pipelines present accuracy results worst than all trials we run in our pipeline, even considering the same conditions of descriptors and leaf size, e.g., 1 cm of leaf size in Boost $US_{0.01}$ trial. When we consider time processing, the difference is even more discrepant when our approach presents in the best case, a frame-rate of 14.18 against 0.09 FPS on the best standard approaches, which represents a remarkable improvement of more than $150\times$ in speed. When using the EfficientNet/FGR pair, our proposal presents AUC (0.4123) three times higher than the Boosted pipeline (0.1372). We did not run the Baseline $US_{0.01}$ because this method is very time-consuming and does not represent a reasonable choice regarding the boosted version (Boost $US_{0.01}$). We found a frame rate of $0.0005$ for a small set of frames experimentally. Besides, the boosted pipeline \cite{marcon2019boosting} gains on accuracy and time performance regarding the traditional version, as seen on the trials with a leaf size of $0.02$ (Baseline $US_{0.02}$ and Boost $US_{0.02}$), and such behavior is also expected on a smaller leaf size.

\vspace{2mm}

\subsubsection{Time processing evaluation}

Now we report the processing rate regarding executing the three stages of our proposed pipeline. \autoref{tab:comparison-fast-ransac} presents the frame processing rate based on a single target object scenario.  We evaluate referring to the first stage execution (\textit{Color only}), the early two stages (Columns \textit{RANSAC}, and \textit{FGR}), and a pipeline's full execution (\textit{+ICP}).

\begin{table*}[htb]
	\centering
	\small
	\vspace{2.5mm}
	\renewcommand{\arraystretch}{1.3}
	\caption{Single target pose estimation FPS. \textit{Color only} refers to object classification, other columns refer to the pose aligment step, coarse (RANSAC and FGR) or fine (plus ICP). \label{tab:comparison-fast-ransac}}
\begin{tabular}{lccccc}
\hline
 & {Color only} & {RANSAC} & {FGR} & {RANSAC + ICP} & {FGR + ICP} \\ \hline
MobileNet v2 \cite{sandler2018mobilenetv2} & {89.49} & 1.89 & 13.89 & 1.82 & {13.57} \\
ResNet101 \cite{he2016resnet} & 52.45 & 1.96 & 13.83 & 1.81 & 13.39 \\
ResNeXt101 32x8d \cite{xie2017resnext} & 33.73 & {2.03} & {14.18} & {2.09} & 13.32 \\
EfficientNet-B7 \cite{tan2019efficientnet} & 22.51 & 1.43 & 8.94 & 1.40 & 8.55 \\ \hline
\end{tabular}
\vspace{2mm}
\end{table*}

At first sight, one can conjecture that a RANSAC-based approach is unpromising when presenting around 2 FPS. However, considering an FGR-based process, the results are indeed encouraging, with 8 FPS for the best accurate method, and more than 13 for the others. For many applications that deal with real-time, a frame rate around eight or more is acceptable. We agree that \textit{the facto} standard for real-time is at least 30 FPS, however, due to the modularity of our proposed pipeline, the stages are independent, and we could use the full execution only to indispensable situations.

\begin{figure}[htb]
	\centering
	\small
	\renewcommand{\arraystretch}{1.2}
	\begin{tabular}{c}
	\includegraphics[clip, trim=1.0cm 0cm 3.6cm 0cm, width=0.95\linewidth]{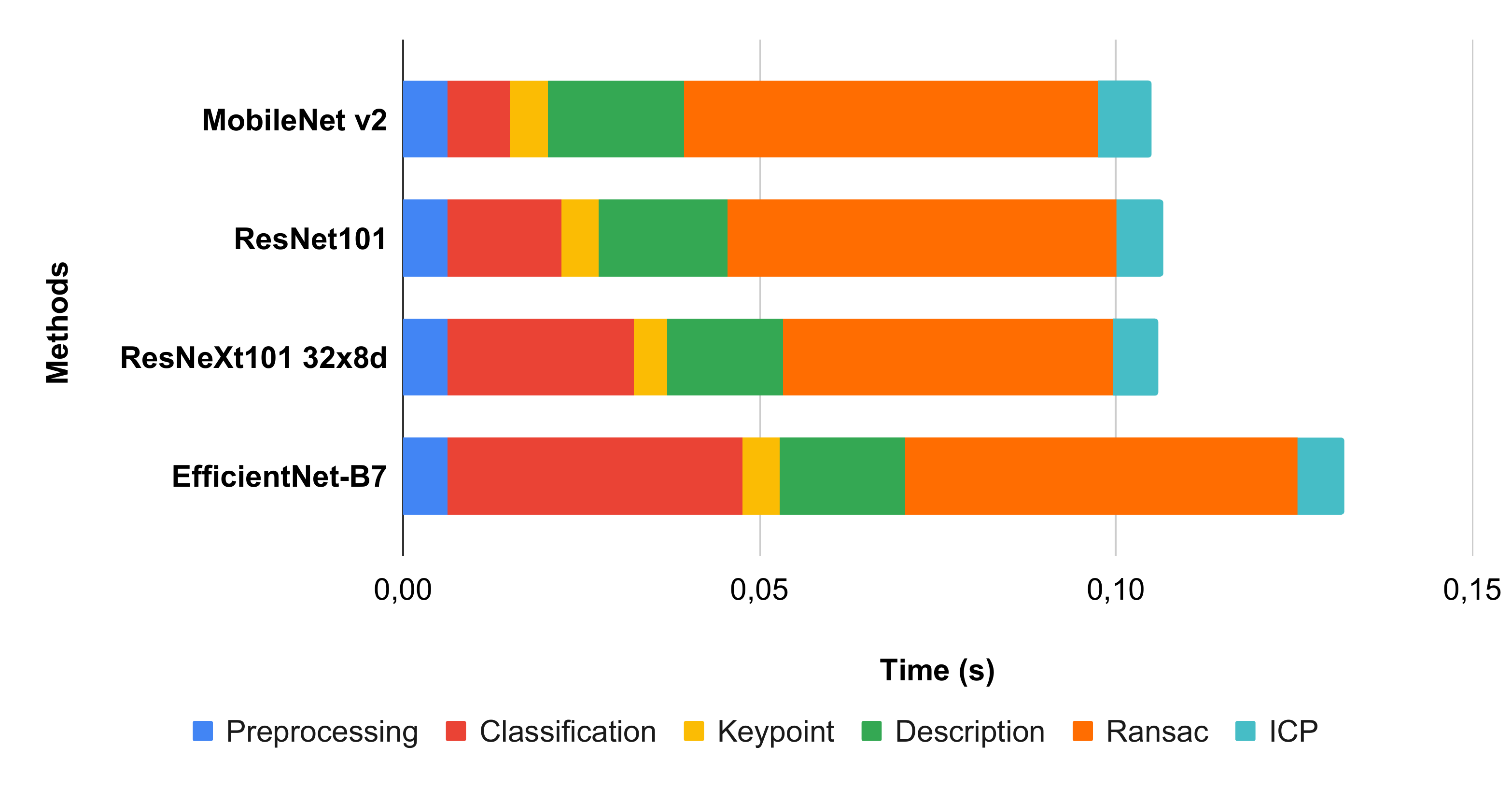} \\
	(a) FGR \\
	\vspace{2mm}
	\includegraphics[clip, trim=1.0cm 0cm 3.6cm 0cm,width=0.95\linewidth]{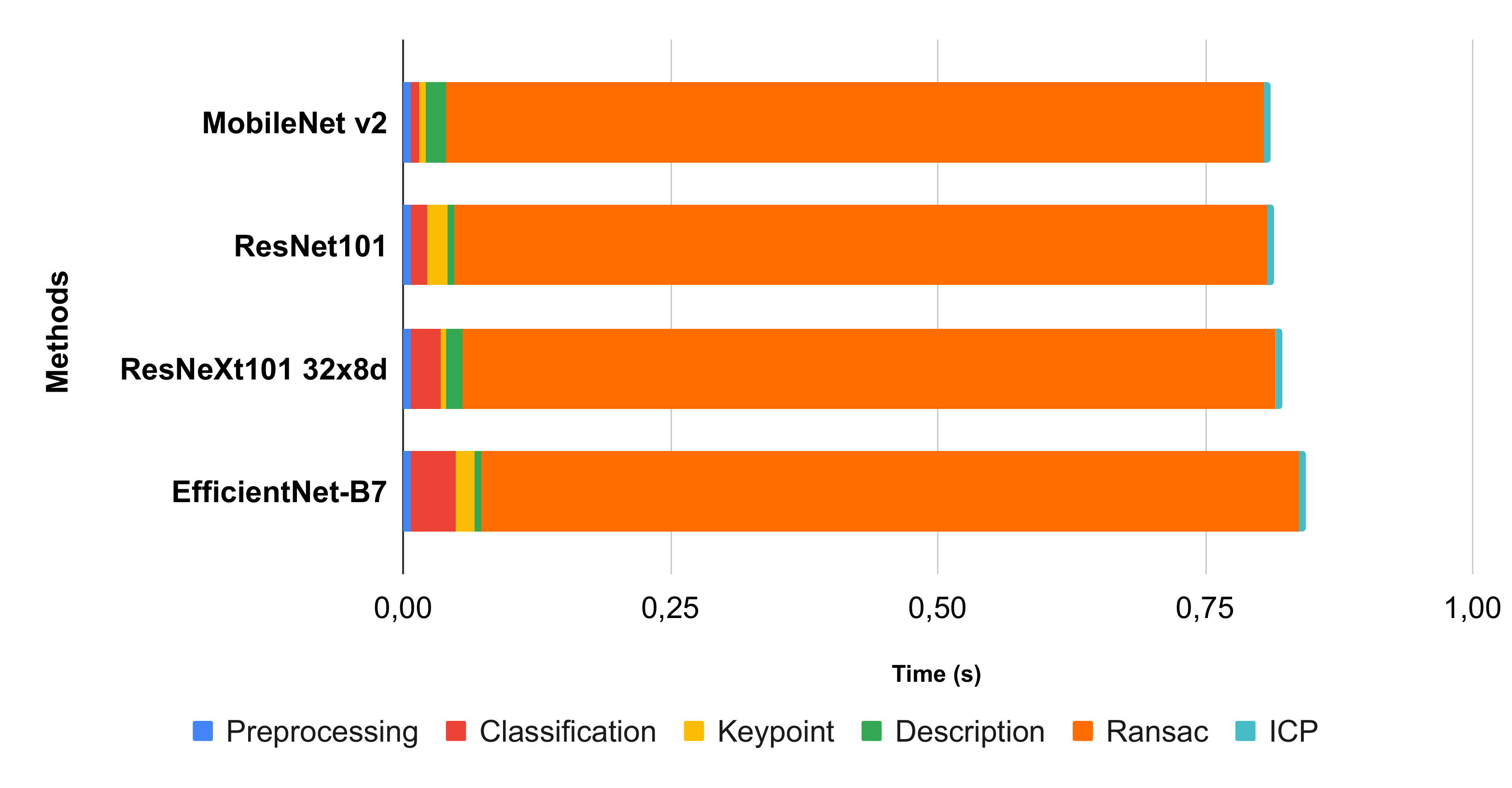} \\
	(b) RANSAC \\
	\end{tabular}
	
	\caption{Processing time (in seconds) of each step of the execution of proposed approach. We consider only successfully detected objects on this comparison. (a) presents times referring to the FGR \cite{zhou2016fast} method, and (b) to RANSAC.}
	\label{fig:processing-time-per-object}
	\vspace{5mm}
\end{figure}

An application scenario may include a target object's location and pose recovering, for instance, by a robot or a visually impaired person. The system could execute a scheduled procedure, localizing this object adopting only the first stage of the pipeline, in real-time. Then, as the subject approaches the objective, we could execute the second stage, estimating a rough transformation, e.g., once a second. Finally, when the object is next to the user, we can run the full pipeline, including the fine-adjustment stage.

To investigate more deeply the processing time of a successfully detected object of our pipeline, we summarize how much time takes each substep in \autoref{fig:processing-time-per-object}. We can infer that two main steps negatively impact the time processing: classification and feature-based estimation. Regarding the former, the correct selection of the network to extract color features is fundamental to speed-up the whole process, presenting a significant difference between the faster \cite{sandler2018mobilenetv2} and the slower \cite{tan2019efficientnet}. We perceive a considerable impact in time processing when using RANSAC instead of FGR for the feature-based stage. In this implementation, we do not use any concurrent processing, which could significantly improve such time for both coarse pose estimation methods. Our pipeline is highly flexible, and the use of recent proposals may enhance our results on coarse estimation, for instance DGR \cite{choy2020deep}.

\subsubsection{Qualitative results}

 \begin{figure*}[htb]
 	\centering
 	\includegraphics[width=0.95\linewidth]{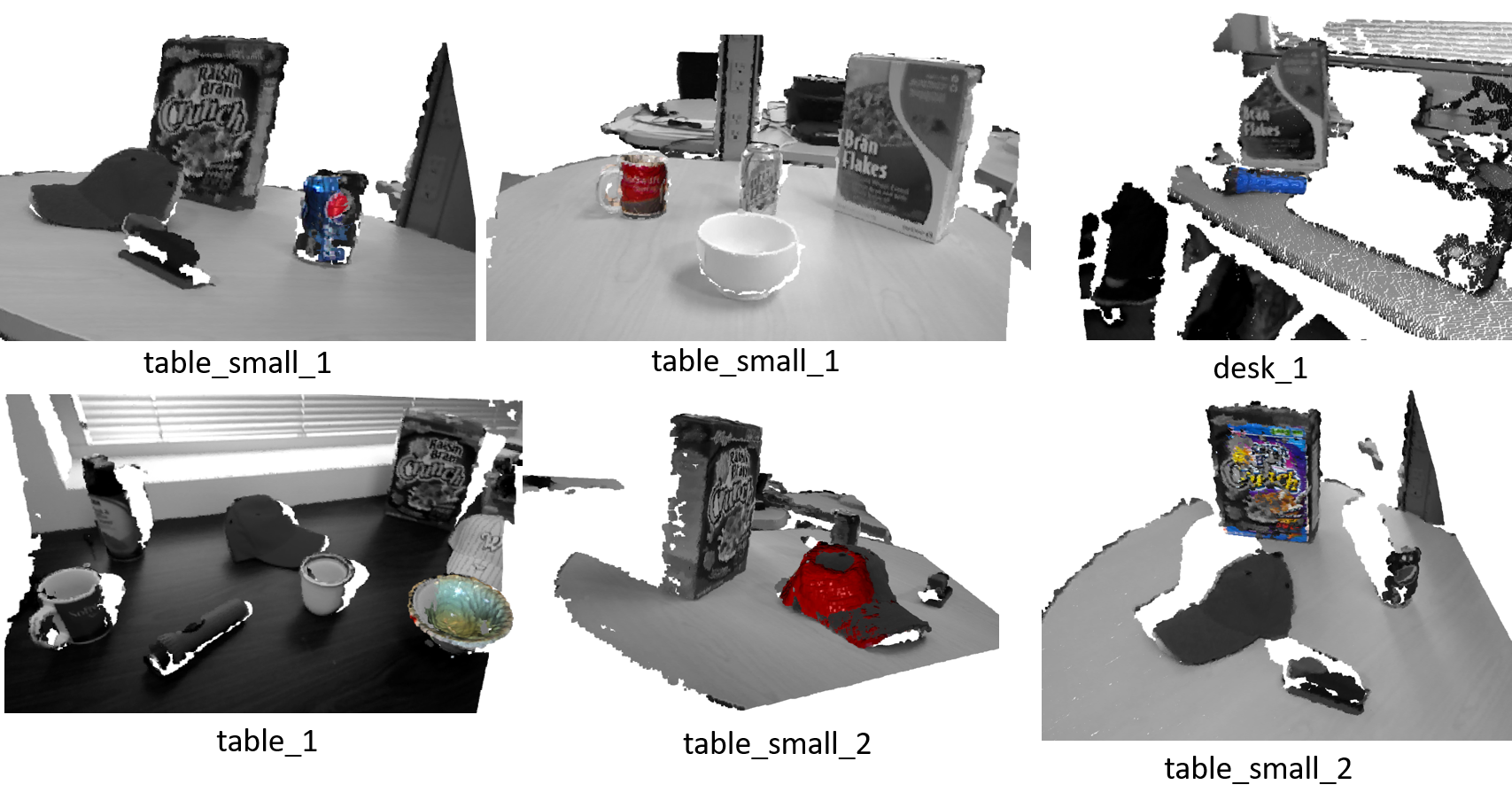}
 	\caption[Qualitative visualizations of successful pose alignment]{Qualitative visualizations of successful pose alignment.}
 	\label{fig:qualitativeposeestimation}
 \end{figure*}

We provide qualitative visualizations of our proposed method (RANSAC + ICP) in \autoref{fig:qualitativeposeestimation}. Our method succeeds in aligning several different shaped models, such as planes (\textit{cereal box}), cylinders (\textit{soda can}, \textit{coffee mugs}, and \textit{flashlights}), and free form models (\textit{caps}). As we perform a rigid transformation to align objects and scenes, the model's choice is fundamental. Examples like the \textit{red cap} that present a crumple on top harm the alignment estimation.  Otherwise, we confirm the robustness of the combination of coarse and fine alignments on the bowl object (bottom row, on the left), partially cropped on the scene cloud. Still, our method infers the pose correctly.

\begin{figure*}[htb]
 	\centering
 	\includegraphics[width=0.95\linewidth]{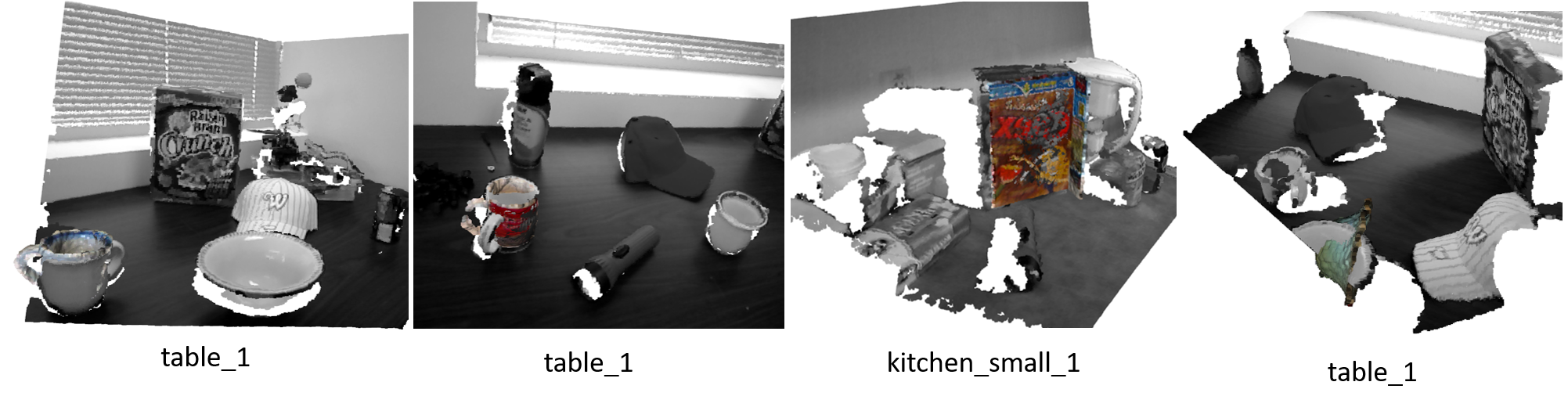}
 	\caption[Qualitative visualizations of wrong pose alignment]{Qualitative visualizations of wrong pose alignment. From left to right: two examples of coffee mugs with a misoriented handles, flipped cereal box, and a rotated bowl.}
 	\label{fig:wrongposeestimation}
 \end{figure*}

In \autoref{fig:wrongposeestimation}, we present some wrong alignments of our proposals. We can observe that the objects' main shape weights a lot on the alignment results. For instance, the \textit{mugs} had the body well aligned but a misalignment on the handle. We also perceive a flip on the cereal box because of the large plane at the front. The \textit{bowl} in the rightmost example fails in aligning, though, different from the previous figure, where the method robustly handled a partial view of a \textit{bowl}, this particular case, have about 50\% only of the object visible. The ICP algorithm estimates a locally minimal transformation, and such misalignments may occur because of inaccurate inputs produced by RANSAC/FGR methods. We espy three potential solutions: using novel CNN-based estimation methods, e.g., DGR \cite{choy2020deep}; adopting more robust local descriptors to the feature-based registration phase, also considering color-based approaches; increasing the number of selected 2.5D views to enhance pose covering of the scenes' objects. The last two cited solutions may negatively affect time-performance.  Despite the misalignments verified, as we reduce the surface search on the scene cloud, we always have an estimation next or even inside the 3D projection of the 2D bounding box outputted by the detection.



\section{\uppercase{Conclusions}}

3D pose estimation is a challenging task, mainly for real-time applications. Sometimes developers must surrender on the precision, aiming the response time. In this paper, we introduced a novel pipeline that proposes to combine the power of color features extractors deep networks, with a local descriptors pipeline to pose estimation in point clouds. We evaluated the detection of objects and achieved almost 83\% on an instance situation, in the best case. This precision is also accompanied by a high frame processing rate, arriving up to 90 FPS. The pose estimation rate is plausible for some applications, and by scheduling the stages of our pipeline, we can reach standard real-time processing. We show experimentally massive improvements concerning accuracy and time processing compared to standard approaches for object recognition and pose estimation. Our approach is $3\times$ more efficient and $150 \times$ faster than traditional and grounded methodologies.

Our three-staged detachable pipeline can be used according to the user/application needs: the color feature classification provides object detection in real-time; the feature-based registration estimates an imprecise but sometimes efficient pose of the scenes' object; the third stage performs a fine alignment of the estimation, resulting in a more accurate result. We believe that our proposal's adoption may help researchers and the industry develop reliable and time-efficient solutions for scene recognition problems from RGB-D data.

Parallelization strategies can improve time results even more and also different local descriptors and keypoint extractors could support this. Findings on the deepnets architectures can help developing an integrated region proposal and object detection algorithm, and state-of-the-art deep learning methods such as SSD \cite{liu2016ssd}, YOLO \cite{redmon2016yolo}, and EfficientDet \cite{tan2020efficientdet} enable such potentiality.

\section{\uppercase{Acknowledgments}}

We would like to thank UTFPR-DV for partly supporting this research work

\bibliographystyle{apalike}
{\small
\bibliography{references}}



\end{document}